\documentclass{article}
\usepackage{iclr2027_conference,times}
\usepackage[T1]{fontenc}

\usepackage{amsmath,amsfonts,bm}

\def\eqref#1{equation~\ref{#1}}

\def\1{\bm{1}}

\DeclareMathAlphabet{\mathsfit}{\encodingdefault}{\sfdefault}{m}{sl}
\SetMathAlphabet{\mathsfit}{bold}{\encodingdefault}{\sfdefault}{bx}{n}

\usepackage{hyperref}
\hypersetup{hidelinks,pdfauthor={Zhuo Chen, Bin Zou, Fanyue Wei, Lixin Duan, Angela Yao, Wen Li}}
\usepackage{url}
\usepackage{booktabs}
\usepackage{capt-of}
\usepackage{wrapfig}
\usepackage{needspace}
\usepackage{multirow}
\usepackage{amsmath}
\usepackage{amssymb}
\usepackage{graphicx}
\usepackage{xcolor}
\definecolor{projectlinkblue}{RGB}{40,120,181}
\usepackage{xspace}

\newcommand{\methodname}{TGuide\xspace}

\newcommand{\FFFF}{\texttt{FFFF}\xspace}

\newcommand{\FPPF}{\texttt{FPPF}\xspace}
\newcommand{\FFPF}{\texttt{FFPF}\xspace}
\newcommand{\FPFF}{\texttt{FPFF}\xspace}

\newcommand{\FRRF}{\texttt{FRRF}\xspace}

\newcommand{\stepF}{\texttt{F}\xspace}
\newcommand{\stepP}{\texttt{P}\xspace}
\newcommand{\stepR}{\texttt{R}\xspace}

\title{\methodname{}: Faster Autoregressive Video Diffusion via Historical Trajectory Guidance}

\author{\begin{minipage}{\dimexpr\textwidth-2\tabcolsep\relax}
\centering
\vspace{8pt}
Zhuo Chen$^{1}$\quad Bin Zou$^{2}$\quad Fanyue Wei$^{3}$\quad Lixin Duan$^{1}$\quad Angela Yao$^{3}$\quad Wen Li$^{1,\dagger}$\\[6pt]
\normalfont\small
$^{1}$University of Electronic Science and Technology of China\\
$^{2}$University of Hong Kong\\
$^{3}$National University of Singapore\\[3pt]
$^{\dagger}$Corresponding author.\\[4pt]
Project Page: \href{https://zhuochen02.github.io/tguide/}{\textcolor{projectlinkblue}{\textbf{https://zhuochen02.github.io/tguide/}}}
\end{minipage}}

\iclrfinalcopy

\begin{document}

\addtocontents{toc}{\protect\setcounter{tocdepth}{-1}}

\maketitle
\lhead{Preprint}

\vspace{-18pt}
\noindent\begin{minipage}{\textwidth}
\centering
\includegraphics[width=\linewidth]{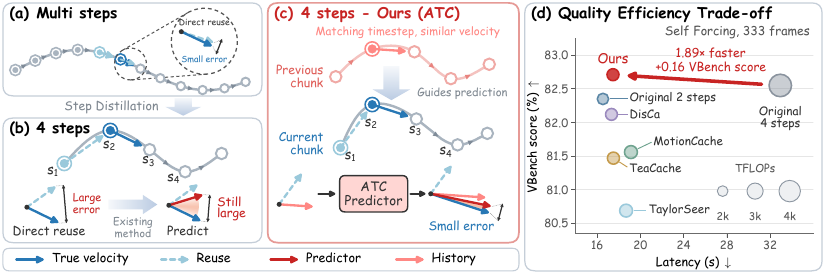}
\captionof{figure}{\textbf{Motivation and overview of \methodname{}.}
Curves and arrows schematically illustrate sampling trajectories and velocity relationships.
\textbf{(a)} With many denoising steps, nearby velocities can be similar, making direct reuse a useful approximation.
\textbf{(b)} Step distillation increases the discrepancy between adjacent step velocities, leading to large errors in both direct reuse and existing prediction methods.
\textbf{(c)} Past sampling trajectories guide current velocity prediction: ATC combines the previous chunk's velocity at the matching timestep with the current chunk's velocity from an earlier step to improve prediction accuracy in 4-step generation.
\textbf{(d)} On Self Forcing with 333-frame generation, our \methodname{} achieves a $1.89\times$ speedup over the original 4-step baseline while improving the VBench score by $0.16$ points. Bubble sizes indicate computational cost in TFLOPs.}
\label{fig:trajectory}
\end{minipage}\par

\vspace{6pt}

\begin{abstract}
\vspace{-10pt}
Can autoregressive video diffusion be accelerated further when only four denoising steps remain? 
Existing feature caching approaches struggle in this regime, as large feature changes between denoising timesteps undermine both direct reuse and lightweight predictors. 
We observe that features across video chunks remain strongly correlated at matching denoising timesteps, offering a complementary source of guidance. 
Building on this insight, we introduce \methodname{}, a framework that accelerates autoregressive video diffusion through a lightweight Anchor-Transport-Correct (ATC) predictor guided by historical sampling trajectories.
Our ATC combines two complementary references: the current chunk’s preceding-step features preserve its spatial structure, while the previous chunk’s same-timestep features provide information at the target denoising stage. 
By integrating these references through spatial retrieval and adaptive correction, ATC improves velocity prediction accuracy despite the large gaps between denoising steps.
To mitigate error accumulation across both denoising steps and video chunks, we train the predictor on its own autoregressive rollouts, allowing it to adapt to states shaped by its own earlier predictions. 
An optional confidence-aware scheduler further adjusts the quality–efficiency trade-off.
Experiments on Self Forcing, Causal Forcing, and HY-WorldPlay show that \methodname{} preserves generation quality while achieving denoising speedups of up to $1.89\times$ over the original 4-step generators.
On Self Forcing, \methodname{} generalizes from 81-frame training videos to 333-frame generation, approximately four times the training length, while exceeding the original model's VBench score.
\end{abstract}

\section{Introduction}
\label{sec:intro}

Autoregressive video diffusion enables continuous video generation by synthesizing successive frames or chunks conditioned on previously generated content~\citep{chen2024diffusionforcing,huang2025selfforcing,yin2025slow}. 
Its sequential generation process allows user controls to guide subsequent content, supporting interactive applications and game simulation~\citep{he2025matrix,sun2025worldplay}. 
To reduce the computational overhead of autoregressive generation, recent methods, including CausVid~\citep{yin2025slow}, Self Forcing~\citep{huang2025selfforcing}, and Causal Forcing~\citep{zhu2026causal}, use distribution matching distillation (DMD) to reduce generation to just four denoising steps per chunk. 
However, each step still requires an expensive Diffusion Transformer (DiT) evaluation to predict the denoising velocity, making it difficult to satisfy the requirements of real-time generation. 
For example, on an NVIDIA A100 GPU, 4-step Self Forcing requires $32.9$ seconds of denoising to generate a $20.8$-second video (333 frames at $16$ FPS; Tab.~\ref{tab:sf-main}).
This shortcoming motivates further acceleration of few-step autoregressive video generation.

\begin{figure}[t]
\centering
\includegraphics[width=\linewidth]{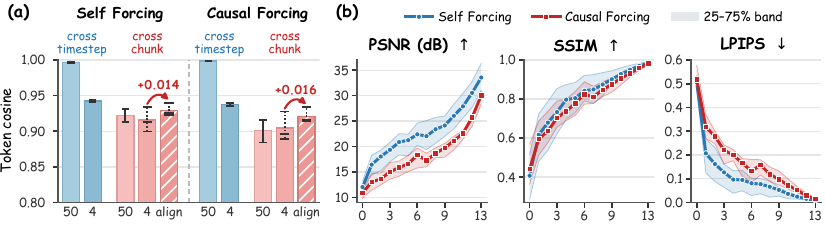}
\vspace{-20pt}
\caption{\textbf{(a) Correlation along denoising time and video time.}
The 4-step autoregressive generators exhibit lower cosine similarity of final hidden states across denoising timesteps than the 50-step teacher references, while correlations across video chunks remain largely preserved.
Optical-flow alignment further improves cross-chunk similarity.
\textbf{(b) Global impact of local approximation.}
Velocity reuse in a single chunk generally causes larger full-video deviations from a full-compute reference when applied earlier in the rollout, motivating more conservative approximation decisions for earlier chunks.
The horizontal axis indicates the index of the chunk perturbed by velocity reuse.
See Appendix~\ref{app:motivation-diagnostics} for diagnostic protocols.
}
\label{fig:boundary}
\label{fig:reuse-chunk}
\vspace{-10pt}
\end{figure}

Complementary to reducing the sampling steps, feature caching methods such as TeaCache~\citep{liu2025timestep} exploit cross-timestep feature redundancy to reduce computation within each step, achieving substantial speedups under multi-step sampling schedules.
However, 4-step sampling introduces larger gaps between adjacent denoising states, weakening the similarity that supports feature reuse and increasing the error of directly reusing cached features (Figs.~\ref{fig:trajectory}(a,b) and~\ref{fig:boundary}(a)).
To mitigate reuse errors, DisCa~\citep{zou2026disca} learns a lightweight predictor to approximate target features from cached features, enabling effective caching in models distilled to $10$--$20$ denoising steps.
However, extending this approach to 4-step generation remains challenging: the predictor must approximate a large change in denoising stage with only a few DiT blocks, leading to reduced prediction accuracy (Fig.~\ref{fig:trajectory}(b)).
Motivated by feature correlations across video frames, C$^3$ache~\citep{zhao2026c3acheacceleratingworldaction} and X-Cache~\citep{zeng2026x} exploit redundancy along the video-time axis of autoregressive DiTs by reusing features from historical chunks.
However, direct reuse across chunks does not account for motion-induced feature differences, and larger motion can lead to greater reuse errors.
Together, these limitations raise a question: \emph{can historical features still provide useful guidance for further acceleration, even when only four denoising steps remain?}

\label{sec:intro:motivation:atc}
To investigate this possibility, we first examine feature correlations along both denoising time and video time using DiT final hidden states.
As shown in Fig.~\ref{fig:boundary}(a), compared with the $50$-step teacher references, the 4-step autoregressive generators exhibit substantially lower feature similarity across denoising timesteps, while similarity across video chunks at matching timesteps remains largely preserved.
Moreover, oracle optical-flow alignment further increases cross-chunk similarity, suggesting that spatial displacement obscures useful historical correspondence.
These observations suggest that historical features can remain informative even when they are unsuitable for direct reuse.
In particular, the previous chunk provides a reference from its sampling trajectory at the target denoising stage, complementing the current chunk's preceding-step information tied to its current content (Fig.~\ref{fig:trajectory}(c)).
This motivates combining these complementary references to improve lightweight feature prediction in few-step autoregressive video generation.

\label{sec:intro:motivation:conf}
Building on these insights, we introduce \methodname{}, a framework for accelerating few-step autoregressive video diffusion through historical trajectory guidance.
At its core is an Anchor-Transport-Correct (ATC) predictor that uses the previous chunk's sampling trajectory to guide current velocity prediction.
\emph{Anchor} forms an initial prediction from the current noisy input and the current chunk's preceding-step features.
To account for spatial displacement across chunks, \emph{Transport} retrieves relevant information from the previous chunk's trajectory at the matching denoising timestep.
\emph{Correct} then selectively incorporates this information to refine the anchor prediction.
ATC operates in DiT final hidden-state space, and the predicted state is decoded into velocity through the frozen output head.
During autoregressive generation, approximate predictions can influence the states encountered in subsequent steps and chunks.
We therefore train ATC through self-rollout on-policy distillation, allowing it to adapt to inputs shaped by its own earlier predictions.
As an optional extension, a confidence-aware scheduler provides additional control over the quality--efficiency trade-off.

Experiments on Self Forcing, Causal Forcing, and HY-WorldPlay demonstrate that \methodname{} accelerates autoregressive video generation while preserving generation quality.
Notably, despite training only on 81-frame videos, \methodname{} achieves a $1.89\times$ denoising speedup on 333-frame Self Forcing generation, approximately four times the training length, while surpassing the original 4-step model's VBench Score (Fig.~\ref{fig:trajectory}(d)).
Ablations further support the effectiveness of historical feature retrieval, correction, and self-rollout training.
The contributions are summarized as follows:
\begin{itemize}
\item We propose \methodname{}, a framework for accelerating few-step autoregressive video diffusion through historical trajectory guidance.
Its Anchor-Transport-Correct (ATC) predictor uses historical sampling trajectories to improve current velocity prediction through spatial retrieval and adaptive correction.
\item We combine self-rollout on-policy distillation with an optional confidence-aware scheduler to support reliable autoregressive acceleration.
\item Experiments across three autoregressive video generators demonstrate favorable quality--efficiency trade-offs and generalization to videos longer than those used for training. Ablations further support the effectiveness of our key designs.
\end{itemize}

\section{Related Work}
\label{sec:related}

\paragraph{Autoregressive Video Diffusion.}
\label{sec:related:ar}

Autoregressive video diffusion enables continuous generation by synthesizing frames or chunks conditioned on historical context.
Diffusion Forcing~\citep{chen2024diffusionforcing} introduces flexible temporal noise modeling, subsequently extended to scalable chunk-wise video generation~\citep{chen2025skyreels,ai2025magi1autoregressivevideogeneration}.
CausVid, Self Forcing, and Causal Forcing advance few-step generation through distillation and improved training--inference alignment~\citep{yin2025slow,huang2025selfforcing,zhu2026causal}, alongside complementary developments in adversarial training, reward-guided optimization, and temporal dependency modeling~\citep{lin2026autoregressive,zhang2026reward,lu2026reward,lu2026raven,liu2026streaming}.
To sustain generation over longer horizons, LongLive~\citep{yang2025longlive} and related approaches improve context management and rollout training to mitigate temporal degradation~\citep{zhang2025framecontextpackingdrift,liu2026rolling,cui2026self,chen2026context,yuan2026helios}.
HY-WorldPlay~\citep{sun2025worldplay} demonstrates action-controlled interactive generation, a capability also explored by recent world models~\citep{wang2026matrix,gao2026infinite}.
Building on few-step autoregressive generators, our work reduces denoising computation through lightweight feature prediction guided by historical sampling trajectories.

\paragraph{Feature Caching for Diffusion Transformers.}
\label{sec:related:cache}

Feature caching accelerates diffusion transformers by reusing intermediate features across denoising timesteps along a sampling trajectory, with fixed or adaptive update schedules~\citep{chen2024deltadittrainingfreeaccelerationmethod,selvaraju2024fora,zhao2025real,zou2024accelerating,liu2025timestep,kahatapitiya2025adaptive,ma2026magcache,bu2026dicache}, while forecasting and learning-based methods improve approximation through extrapolation, optimized caching policies, and learned feature transformations~\citep{liu2025reusing,zheng2025forecastcalibratefeaturecaching,liu2025freqca,zheng2026let,ma2024learning,huang2024harmonica,liu2026lincaacceleratingdiffusionmodels}.
DisCa~\citep{zou2026disca} extends this direction with a lightweight predictor conditioned on cached features and the current noisy input, enabling caching in step-distilled models.
These approaches primarily exploit information within a single sampling trajectory.
However, under 4-step sampling, the larger gaps between adjacent denoising states make both direct reuse and lightweight prediction along this trajectory more challenging.
For autoregressive generation, recent methods adapt caching to chunk-dependent dynamics~\citep{ma2026flow,xu2026motion,nan2026accelerating}.
C$^3$ache~\citep{zhao2026c3acheacceleratingworldaction} reuses the aggregate DiT residual across inference chunks at matching denoising steps, with scheduled cache refreshes.
X-Cache~\citep{zeng2026x} performs block-level residual reuse with adaptive gating.
Both methods directly reuse residuals from historical trajectories at corresponding denoising stages.
However, matching denoising stages does not guarantee spatial correspondence: motion can displace historical content, while direct residual reuse does not explicitly recover these correspondences.
In contrast, our ATC predictor retrieves and adapts information from historical sampling trajectories at the target denoising stage to guide current velocity prediction.
This allows historical features to provide useful guidance even when direct reuse is unreliable, improving prediction accuracy at low computational cost.

\section{Method}
\label{sec:method}

\subsection{Preliminaries}
\label{sec:preliminaries}

\noindent\textbf{Autoregressive Video Diffusion.} Videos are generated along two axes: video chunks $j$ and denoising steps $i$ within each chunk.
Let $x_i^j$ denote the noisy latent at step $i$ of chunk $j$, and let $\mathcal{C}^{j}$ collect the external conditions and historical KV cache from earlier chunks.
Starting from noise, each chunk is refined over $N$ denoising steps at timesteps $t_0,\ldots,t_{N-1}$ conditioned on $\mathcal{C}^{j}$.
Let $\mathcal S_i$ denote the sampling update at step $i$.
Each step performs the update
\begin{equation}
x_{i+1}^{j}
=
\mathcal{S}_i\!\left(
x_i^{j},\, v_\theta(x_i^{j}, t_i, \mathcal{C}^{j})
\right).
\label{eq:base_sampling}
\end{equation}
Here, $v_\theta(\cdot)$ is the velocity prediction from the Diffusion Transformer (DiT) with parameters $\theta$.

\paragraph{Few Step Distillation.} Generating $K$ chunks requires $K\cdot N$ full DiT denoising evaluations.
Step distillation trains a student generator to approximate a teacher's generation process using fewer denoising steps, thereby reducing $N$.
This work considers already-distilled autoregressive generators, with $N=4$; each of the four denoising steps ordinarily still requires a full DiT evaluation.

\paragraph{Feature Caching.} One can also limit the computational cost of select DiT evaluations.
Let $\mathcal{H}_i^{j}$ denote cached representations available at step $i$ of chunk $j$.
At selected steps, a reuse or prediction function $\mathcal{G}$ supplies an approximate velocity $\hat v_i^j$ in place of $v_\theta$ in Eq.~\ref{eq:base_sampling}:
\begin{equation} 
\hat v_i^{j}
=
\mathcal{G}\!\left(
x_i^{j},t_i,\mathcal{C}^{j};\mathcal{H}_i^{j}
\right).
\label{eq:cache_approximation}
\end{equation}
Cached computation can approximate the target features through direct reuse, extrapolation, or a lightweight learned predictor. In few-step generators, larger gaps between denoising states make these cross step approximations less accurate, motivating more informative references.

\subsection{Problem Setup and Motivation}
\label{sec:method:problem}

We seek to accelerate the 4-step generator by replacing selected DiT evaluations with lightweight predictions.
We first consider reusing computation across
denoising steps within the same chunk, and then show how preceding chunks provide a complementary source of information.

\paragraph{Prediction from cached computation.}
Following Eq.~\ref{eq:cache_approximation}, we use the cached velocity from an earlier denoising step of the current chunk as an anchor for prediction:
\begin{equation}
\hat v_{i,P}^{j}
=
\mathcal{P}\!\left(
x_i^{j}, t_i, \mathcal{C}^{j};
v_{i_a}^{j}
\right).
\label{eq:within_chunk_prediction}
\end{equation}
Here $i_a<i$ denotes the anchor step, usually
$i_a=i-1$. Its cached velocity $v_{i_a}^{j}$ may come from a full DiT evaluation or an accepted prediction.   Eq.~\ref{eq:within_chunk_prediction} specializes the generic cache $\mathcal{H}_i^{j}$ in Eq.~\ref{eq:cache_approximation} to this reference. It is a natural starting point because the cached output and the target correspond to the same video content. However, with only four denoising steps, the cached output $v_{i_a}^{j}$ may differ substantially from the target output $v_i^{j}$ because their timesteps are widely separated, consistent with the lower cross timestep feature similarity in Fig.~\ref{fig:boundary}(a). Consequently, the predictor bridges the denoising gap with substantially less computation than the full DiT, compromising prediction accuracy as illustrated in Fig.~\ref{fig:trajectory}(b).

\paragraph{Guidance from the previous chunk.}
Our key observation is that, under 4-step autoregressive generation, feature similarity across chunks at matching timesteps remains comparable to that across denoising steps within a chunk (see Fig.~\ref{fig:boundary}(a)).
We therefore augment the predictor with $v_i^{j-1}$, the cached velocity used at the target timestep of the previous chunk, to guide the current prediction as illustrated in Fig.~\ref{fig:trajectory}(c):
\begin{equation}
\hat v_{i,\mathrm{hist}}^{j}
=
\mathcal{P}_{\mathrm{hist}}\!\left(
x_i^{j}, t_i, \mathcal{C}^{j};
v_{i_a}^{j}, v_i^{j-1}
\right).
\label{eq:history_prediction}
\end{equation}
Since the preceding chunk has already been processed, $v_i^{j-1}$ is available during causal inference, whether it came from a full DiT evaluation or an accepted prediction. However, the previous chunk's output $v_i^{j-1}$ cannot simply be copied, because motion and viewpoint changes may displace content or reveal new regions. Conversely, the current chunk's anchor output $v_{i_a}^{j}$ reflects its spatial structure but belongs to an earlier denoising stage. The two references therefore provide complementary information, motivating a predictor that adapts historical content to the current chunk while accounting for the change in denoising stage.

\begin{figure}[t]
\centering
\includegraphics[width=\linewidth]{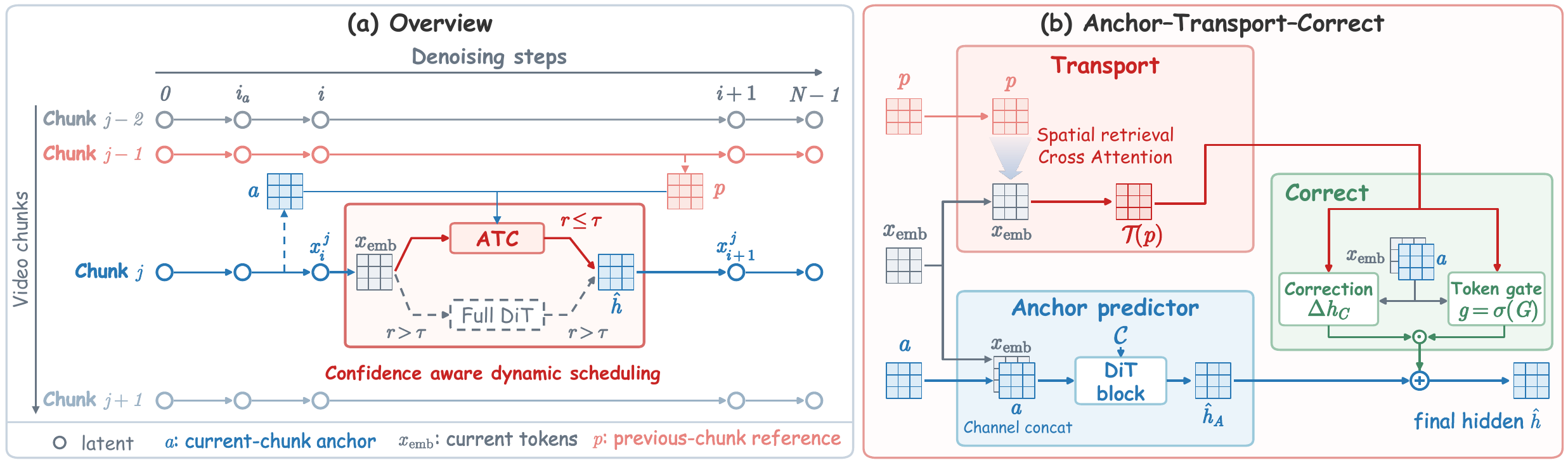}
\vspace{-20pt}
\caption{\textbf{Overview of \methodname{}.}
\textbf{(a)} ATC replaces selected full DiT forward passes by predicting final hidden states from current tokens $x_{\mathrm{emb}}$, the current chunk's anchor $a=h_{i_a}^j$, and the previous chunk's reference $p=h_i^{j-1}$ at the same timestep.
Optional confidence-aware scheduling accepts predictions when $r\leq\tau$ and otherwise invokes the full DiT.
\textbf{(b)} The anchor predictor estimates the current hidden state, Transport retrieves relevant historical features through cross-attention, and Correct applies a token-wise gated correction to produce $\hat h$.}
\label{fig:pipeline}
\end{figure}

\paragraph{Method overview.}
As illustrated in Fig.~\ref{fig:pipeline}, \methodname{} uses an Anchor-Transport-Correct (ATC) predictor to combine the current chunk's denoising anchor with historical information at the target timestep (Sec.~\ref{sec:method:atc}).
The ATC predictor first evolves the current chunk anchor towards the target denoising stage.
It then retrieves relevant historical information across spatial positions, rather than assuming fixed position correspondence, and uses a token-wise gate to control the resulting correction.
This decomposition lets historical computation refine the prediction where correspondence is useful, while preserving the anchor-based estimate where historical information is unreliable.
An optional confidence-aware scheduler determines whether to accept each prediction or invoke the full DiT (Sec.~\ref{sec:method:conf}), while training on self-rollouts prepares the predictor to consume its own approximations (Sec.~\ref{sec:method:train}).

\subsection{Anchor-Transport-Correct predictor}
\label{sec:method:atc}
\label{sec:method:motivation:prev}

We now describe the ATC architecture for the history-guided predictor in Eq.~\ref{eq:history_prediction}.
Although the formulation is expressed in velocity space, ATC predicts the DiT's final hidden state $\hat h$ in practice, which is then mapped to velocity through the frozen output head: $\hat v=\mathrm{Head}(\hat h,t_i)$.
For brevity, let $x_{\mathrm{emb}}$ denote the current noisy latent $x_i^j$ after the input head. We write $a=h^j_{i_a}$ for the current chunk's cached anchor, $p=h_i^{j-1}$ for the previous chunk's final hidden state at the same timestep, and $C=C^j$ for the current chunk's conditions and historical KV context.
During inference, both $a$ and $p$ may come from either full evaluations or earlier accepted predictions.
Historical KV and generation conditions are supplied to the predictor block as in the base model.

\noindent\textbf{Anchor: evolve the current chunk along denoising steps.}
The cached anchor $a$ retains the spatial layout of the current content, but represents an earlier denoising stage. We combine it with the current noisy tokens $x_{\mathrm{emb}}$ and the context $\mathcal C$ to predict a denoising increment:
\begin{equation}
\hat h_A
= a + \Delta h_D\big(x_{\mathrm{emb}},\,a,\,\mathcal C\big).
\label{eq:anchor}
\end{equation}
Here $\Delta h_D(\cdot)$ is implemented with a few DiT blocks, initialized with the weights from the full DiT.

\paragraph{Transport: retrieve history at the matching denoising stage.}
To account for motion and viewpoint changes between chunks, we use current tokens as queries to retrieve spatially corresponding information from the previous chunk's cached state $p$:
\begin{equation}
\mathcal T(p)=\mathrm{CrossAttn}\!\big(Q(x_{\mathrm{emb}}),\,K(p),\,V(p)\big).
\label{eq:transport}
\end{equation}
The transported state $\mathcal T(p)$ contains historical features retrieved for the current tokens.

\paragraph{Correct: selectively refine the anchor evolution.}
The \emph{Correct} branch combines transported history with the current inputs, using a scalar gate per token to control the correction:
\begin{equation}
\begin{aligned}
\hat h
&= \hat h_A + g\odot\Delta h_C\big(x_{\mathrm{emb}},\,a,\,\mathcal T(p)\big)\\
&= a
+ \underbrace{\Delta h_D\big(x_{\mathrm{emb}},\,a,\,\mathcal C\big)}_{\text{denoising evolution}}
+ \underbrace{g\odot\Delta h_C\big(x_{\mathrm{emb}},\,a,\,\mathcal T(p)\big)}_{\text{gated cross chunk correction}}.
\end{aligned}
\label{eq:atc}
\end{equation}
Here, the lightweight MLP $\Delta h_C(\cdot)$ predicts a correction to $\hat h_A$ from the current tokens, anchor, and transported history.
A network $G(\cdot)$ controls its strength through $g=\sigma(G(x_{\mathrm{emb}},a,\mathcal T(p)))$, where $\sigma$ is the sigmoid function.
The operator $\odot$ applies each token's scalar gate across feature channels.
\paragraph{Efficient Architecture Design.}
On Self Forcing, ATC uses a single DiT block and requires only $5.3\%$ of the FLOPs and $6.8\%$ of the latency of a full DiT forward pass.
Unlike TaylorSeer~\citep{liu2025reusing} and X-Cache~\citep{zeng2026x}, which cache intermediate outputs for individual DiT blocks, ATC retains one final hidden state tensor per cached sampling position.
This reduces the additional memory required for feature caching, which is particularly valuable for long video generation, where historical KV caches already consume substantial memory.

\subsection{Confidence aware dynamic scheduling}
\label{sec:method:conf}
\label{sec:method:motivation:chunk}
Fig.~\ref{fig:boundary}(b) shows that introducing velocity reuse in earlier chunks generally leads to larger deviations over the full generated video. This observation motivates a scheduler that combines an estimate of local prediction error with a chunk-position prior, favoring more conservative acceptance decisions earlier in the rollout.
While caching methods such as TeaCache~\citep{liu2025timestep} rely on manually designed proxies to estimate reuse error, such proxies may not fully capture the complex feature dynamics of long-video generation.
We instead attach a lightweight confidence head to ATC, learning to estimate its relative hidden-state error $\hat\epsilon_{j,i}$ from predictor features and  sampling conditions.

We construct a scheduling risk score by weighting the estimated local error with the outgoing denoising span and a chunk-position prior:
\begin{equation}
r_{j,i}=\hat\epsilon_{j,i}\,\frac{|t_i-t_{i+1}|}{T}
\bigl(1+\beta\alpha_j\bigr),
\label{eq:risk}
\end{equation}
where $T$ normalizes the timestep span, $\alpha_j$ decreases linearly from one to zero across predictor-eligible chunks, and $\beta\geq0$ controls the additional weight assigned to earlier chunks.
Given a risk threshold $\tau$, we accept the prediction when $r_{j,i}\leq\tau$ and otherwise invoke the full DiT.
Confidence-head details and prior definitions are provided in Appendix~\ref{app:dynamic-scheduling}.

\subsection{Two stage Training}
\label{sec:method:train}

\paragraph{Stage 1: Offline regression with teacher forcing.}
We first train ATC offline on trajectories collected from the frozen full DiT. Both the anchor hidden states and the previous chunk’s same-timestep hidden states are taken from these trajectories, and ATC is supervised to match the full DiT’s final hidden states and corresponding velocity predictions. This stage provides an initialization for subsequent self-rollout training.

\paragraph{Stage 2: Self-rollout on-policy distillation.}
During inference, however, ATC operates on states affected by its own imperfect predictions. These prediction errors propagate to subsequent noisy latents and cached features, creating a train-test distribution gap along both the denoising and chunk axes. To reduce this mismatch, we follow the self-rollout principle of Self Forcing~\citep{huang2025selfforcing} and perform autoregressive rollouts with ATC during training. Its predictions drive sampling updates and populate the feature cache, while generated chunks provide context for subsequent chunks. We fine-tune ATC on these self-generated trajectories while keeping the base generator frozen.
Detailed training objectives and optimization settings are provided in Appendix~\ref{app:hparams}.

\section{Experiments}
\label{sec:exp}

\subsection{Experimental Setups}
\label{sec:exp:setup}
\label{sec:exp:cache}

\noindent\textbf{Baselines and settings.} We evaluate \methodname{} on three models: Self Forcing, Causal Forcing, and HY-WorldPlay.
We compare with the original 4-step and 2-step models, TeaCache \citep{liu2025timestep}, TaylorSeer~\citep{liu2025reusing}, MotionCache \citep{xu2026motion}, and DisCa \citep{zou2026disca}.
Schedules use \stepF{} for full DiT evaluation, \stepP{} for prediction, and \stepR{} for velocity reuse.
The first chunk and the first denoising step of each subsequent chunk use full DiT.
All experiments use NVIDIA A100 GPUs at $832\times480$ resolution, with \methodname{} using a fixed \FPPF{} schedule in the main comparisons.
Training and baseline implementation details are provided in Appendices~\ref{app:setup-details} and~\ref{app:cache}.

\paragraph{Evaluation protocol.}
\label{sec:exp:cost}
We measure efficiency using TFLOPs, latency, and speedup over the original 4-step model.
We report Quality, Semantic, and Total Scores.
We use the full VBench suite~\citep{huang2024vbench} for the main Self Forcing comparison and the ATC and self-rollout on-policy distillation ablations.
Additional Causal Forcing results and dynamic scheduling ablations are provided in Appendices~\ref{app:sf-dynamic} and~\ref{app:dynamic-scheduling}, respectively.
For camera-conditioned image-to-video generation, we evaluate the generated videos using a subset of the VBench I2V suite.
We follow HY-WorldPlay's camera evaluation protocol, using four cyclic camera trajectories that revisit previously observed views.
PSNR, SSIM, and LPIPS are computed following HY-WorldPlay's protocol, including comparisons against the original 4-step outputs and under the self-comparison protocol, which compares corresponding views along the return and initial paths.
Further evaluation details are provided in Appendix~\ref{app:evaluation}.

\subsection{Qualitative Comparison}
\label{sec:exp:qualitative}

\begin{figure}[t]
\centering
\includegraphics[width=\linewidth]{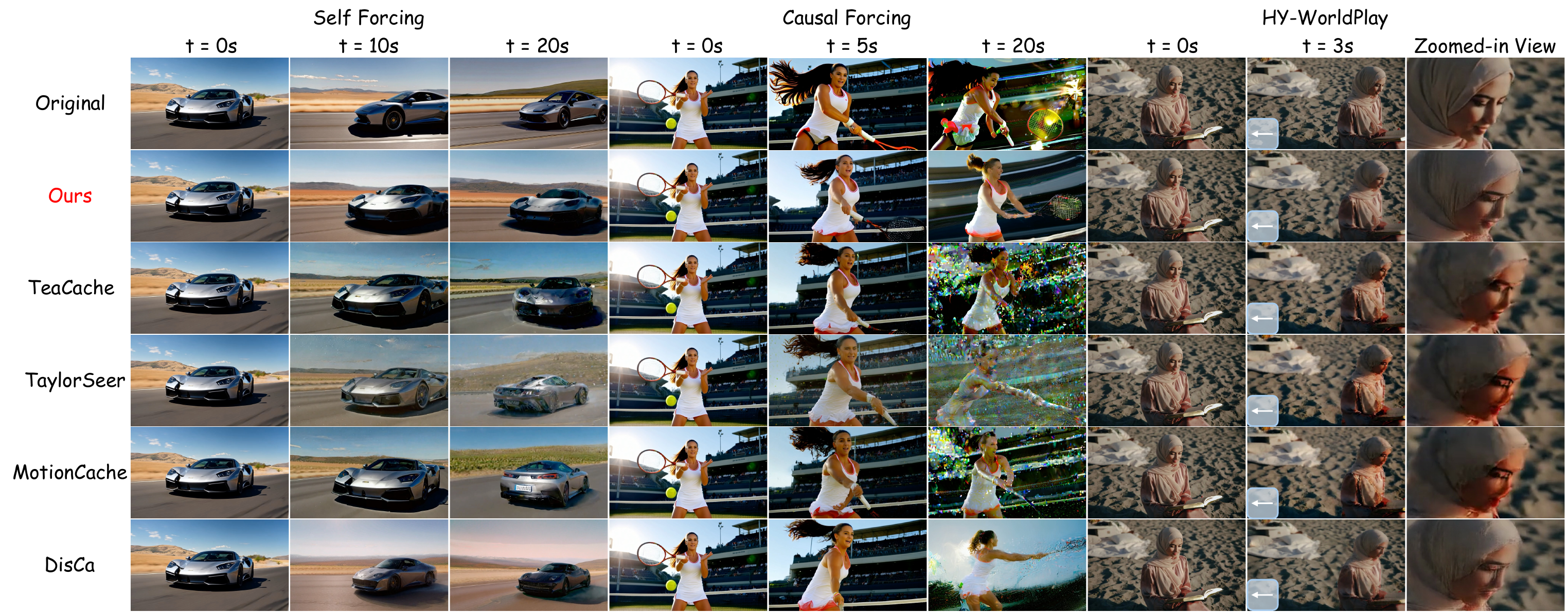}
\vspace{-20pt}
\caption{Qualitative comparisons on Self Forcing, Causal Forcing, and HY-WorldPlay.}
\label{fig:qualitative}
\vspace{-10pt}
\end{figure}

Fig.~\ref{fig:qualitative} illustrates how acceleration affects generated content over time.
In the Self Forcing example, all competing acceleration methods exhibit changes in vehicle appearance or distortions in body structure by $20$ s, whereas \methodname{} better preserves the vehicle's visual identity and structure.
The Causal Forcing example shows more pronounced degradation: several competing methods produce extensive colorful artifacts at $20$ s, while \methodname{} maintains a cleaner depiction of the player and court, with fewer conspicuous artifacts than even the original model in this example.
For HY-WorldPlay, the enlarged views at $3$ s show that \methodname{} better preserves facial contours under camera motion than the competing acceleration methods.
Together, these examples demonstrate improved preservation of subject structure and local detail, supporting visual consistency over extended generation.
Additional visualizations are provided in Appendix~\ref{app:qualitative}.

\subsection{Quantitative Evaluation}
\label{sec:exp:main}

\paragraph{Text-to-video generation.}
\methodname{} preserves generation quality while substantially reducing denoising computation.
On Self Forcing, \methodname{} achieves a Total Score of $84.23$ at a $1.80\times$ speedup for 81-frame generation, closely matching the original 4-step model's $84.28$ (Tab.~\ref{tab:sf-full-vbench}). At comparable latency, it outperforms DisCa by $0.37$ points and the original 2-step model by $0.14$ points, demonstrating a favorable quality--efficiency tradeoff under 4-step sampling. 
Although both training stages use only 81-frame videos, \methodname{} maintains its advantage at $333$ frames, approximately four times the training length.
It achieves a Total Score of $82.71$ at a $1.89\times$ speedup, surpassing DisCa by $0.59$ points at comparable latency and even exceeding the original 4-step model's $82.55$.
These results suggest that historical guidance and self-rollout training help maintain generation quality when generalizing to substantially longer videos.
Detailed 16-dimension Self Forcing scores appear in Tab.~\ref{tab:sf-vbench-dimensions}; Causal Forcing shows a similar advantage (Appendix~\ref{app:causal-forcing}, Tab.~\ref{tab:causal-main}).

\paragraph{Camera-conditioned image-to-video generation.}
\methodname{} achieves a $1.73\times$ speedup while improving both fidelity to the original generator and consistency when revisiting previously generated content (Tab.~\ref{tab:hy-main}).
Against the original 4-step outputs, it achieves the best PSNR, SSIM, and LPIPS among accelerated methods.
It also leads these methods under self-comparison, where frames on the return path are compared with corresponding frames from the initial pass.
These complementary results suggest that the gains extend beyond matching the original outputs to preserving content across camera motion, consistent with our design of spatial retrieval and selective historical correction to adapt previous content to the current view.

\begin{table}[t]
\caption{Performance comparison on Self Forcing using the full VBench benchmark.}
\label{tab:sf-main}
\label{tab:sf-full-vbench}
\label{tab:cache}
\centering
\scriptsize

\setlength{\tabcolsep}{1.5pt}
\begin{tabular*}{\linewidth}{@{\extracolsep{\fill}}cccccc@{\hspace{8pt}}ccccc@{}}
\toprule
\multirow{2}{*}{Method} & \multicolumn{5}{c}{\textbf{81 frames}} & \multicolumn{5}{c}{\textbf{333 frames}} \\
\cmidrule(lr){2-6}\cmidrule(l){7-11}
 & Latency (s) $\downarrow$ & Speedup $\uparrow$ & Total $\uparrow$ & Quality $\uparrow$ & Semantic $\uparrow$ & Latency (s) $\downarrow$ & Speedup $\uparrow$ & Total $\uparrow$ & Quality $\uparrow$ & Semantic $\uparrow$ \\
\midrule
Original 4 steps & 6.53 & 1.00$\times$ & 84.28 & 84.63 & 82.87 & 32.93 & 1.00$\times$ & 82.55 & 82.93 & 81.04 \\
\midrule
Original 2 steps & 3.48 & \textbf{1.88$\times$} & \underline{84.09} & 84.46 & 82.59 & 16.53 & \textbf{1.99$\times$} & \underline{82.35} & 82.76 & 80.71 \\
TeaCache & 3.83 & 1.71$\times$ & 83.91 & 84.14 & 83.01 & 17.50 & 1.88$\times$ & 81.47 & 81.82 & 80.06 \\
TaylorSeer & 3.78 & 1.73$\times$ & 83.68 & 83.93 & 82.69 & 18.67 & 1.76$\times$ & 80.69 & 81.05 & 79.26 \\
MotionCache & 3.96 & 1.65$\times$ & \underline{83.97} & 84.25 & 82.85 & 19.12 & 1.72$\times$ & 81.56 & 81.89 & 80.24 \\
DisCa & 3.59 & \underline{1.82$\times$} & 83.86 & 84.17 & 82.62 & 17.32 & \underline{1.90$\times$} & \underline{82.12} & 82.44 & 80.84 \\
\midrule
Ours & 3.62 & \underline{1.80$\times$} & \textbf{84.23} & 84.59 & 82.80 & 17.46 & \underline{1.89$\times$} & \textbf{82.71} & 83.12 & 81.06 \\
\bottomrule
\end{tabular*}
\vspace{-10pt}
\end{table}

\begin{table}[t]
\caption{Performance comparison on HY-WorldPlay at 125 frames.}
\label{tab:hy-main}
\centering
\scriptsize
\setlength{\tabcolsep}{1.5pt}
\begin{tabular*}{\linewidth}{@{\extracolsep{\fill}}c ccc@{\hspace{6pt}}  ccc@{\hspace{6pt}}ccc@{}}
\toprule
\multirow{2}{*}{Method}
& \multicolumn{3}{c}{Quality and efficiency}
& \multicolumn{3}{c}{vs. Original 4 steps}
& \multicolumn{3}{c}{Self-comparison} \\
\cmidrule(lr){2-4}\cmidrule(lr){5-7}\cmidrule(l){8-10}
& VBench $\uparrow$
& Latency (s) $\downarrow$ & Speedup $\uparrow$
& PSNR $\uparrow$ & SSIM $\uparrow$ & LPIPS $\downarrow$
& PSNR $\uparrow$ & SSIM $\uparrow$ & LPIPS $\downarrow$ \\
\midrule
Original 4 steps
& 84.26
& 49.61 & 1.00$\times$
& --- & --- & ---
& 16.75 & 0.5107 & 0.2820 \\
\midrule
Original 2 steps
& 84.18
& 26.81 & \textbf{1.85}$\times$
& 19.72 & 0.7122 & 0.1480
& 16.68 & 0.5124 & 0.2963 \\
TeaCache
& 84.20
& 30.06 & 1.65$\times$
& 18.63 & 0.7230 & 0.1413
& 16.44 & 0.5088 & 0.3026 \\
TaylorSeer
& \underline{84.21}
& 29.76 & 1.67$\times$
& 18.60 & 0.6845 & 0.1670
& 16.60 & 0.5128 & 0.2983 \\
MotionCache
& \textbf{84.24}
& 30.02 & 1.65$\times$
& 18.59 & 0.6843 & 0.1672
& 16.60 & 0.5135 & 0.2986 \\
DisCa
& 84.17
& 28.39 & \underline{1.75}$\times$
& 19.87 & 0.7293 & 0.1296
& 16.81 & 0.5091 & 0.2955 \\
\midrule
Ours
& \textbf{84.24}
& 28.75 & \underline{1.73}$\times$
& \textbf{20.52} & \textbf{0.7611} & \textbf{0.1144}
& \textbf{16.87} & \textbf{0.5136} & \textbf{0.2872} \\
\bottomrule
\end{tabular*}
\end{table}

\subsection{Ablation Studies}
\label{sec:exp:ablations}

We conduct all ablation studies on Self Forcing.
We first compare predictor architectures under a common offline training protocol, and then evaluate the additional benefit of self-rollout on-policy distillation for both DisCa and ATC.

\paragraph{Anchor-Transport-Correct ablation.}
\label{sec:exp:cross}
\label{sec:exp:atc}
Tab.~\ref{tab:prevcond} evaluates how historical trajectories guide prediction on 81-frame videos.
Here, $p$ is the previous chunk's cached final hidden state at the same denoising timestep (Sec.~\ref{sec:method:atc}).
All learned variants use the same Stage~1 offline regression protocol and fixed \FPPF{} schedule.
\FRRF{} serves as the direct reuse baseline, while \emph{w/o $p$} uses only the anchor predictor with causal KV context.
Replacing direct reuse with learned prediction improves the Total Score from $83.14$ to $83.86$, showing the benefit of adapting cached features to the current denoising state.

\Needspace{13\baselineskip}
\begin{wraptable}{r}{0.52\textwidth}
\vspace{-8pt}
\caption{ATC ablation under Stage 1 training.}
\label{tab:prevcond}
\label{tab:atc}
\centering
\scriptsize
\setlength{\tabcolsep}{1.5pt}
\renewcommand{\arraystretch}{1.05}
\begin{tabular*}{\linewidth}{@{\extracolsep{\fill}}ccccc@{}}
\toprule
Method & Total $\uparrow$ & TFLOPs $\downarrow$ & Latency (s) $\downarrow$ & Speedup $\uparrow$ \\
\midrule
\shortstack[c]{Original 4 steps} & 84.28 & 793 & 6.53 & 1.00$\times$ \\
\midrule
\shortstack[c]{\FRRF{}} & 83.14 & 429 & 3.42 & 1.91$\times$ \\
\shortstack[c]{w/o $p$} & 83.86 & 444 & 3.59 & 1.82$\times$ \\
\shortstack[c]{Concat $p$} & 83.98 & 445 & 3.59 & 1.82$\times$ \\
\shortstack[c]{Anchor + Transport} & 84.05 & 448 & 3.60 & 1.81$\times$ \\
\midrule
ATC & \textbf{84.08} & 452 & 3.62 & 1.80$\times$ \\
\bottomrule
\end{tabular*}
\end{wraptable}

\emph{Concat $p$} directly concatenates historical hidden states, current tokens, and the anchor along the channel dimension at matching spatial positions, improving the score to $83.98$.
\emph{Anchor + Transport} retrieves historical features without the correction module and raises the score to $84.05$, suggesting that spatial retrieval makes history more useful than direct concatenation.
Finally, ATC adds the correction module and reaches $84.08$, supporting the benefit of selectively refining the anchor prediction with retrieved history.
Together, retrieval and correction improve over the predictor without $p$ by $0.22$ points with only $0.03$ s of additional latency, demonstrating the value of adapting historical content beyond the context already supplied by causal KV.

\paragraph{Confidence aware dynamic scheduling.}
\label{sec:exp:conf}
We treat dynamic scheduling as an optional extension and use \FPPF{} in the main experiments to facilitate comparison with fixed strategies such as original 2-step and DisCa.
Dynamic scheduling provides a flexible tradeoff between generation quality and efficiency; a suitable confidence threshold achieves better quality at lower latency than the fixed \FPFF{} and \FFPF{} schedules.
Additional comparisons and detailed scheduling ablations are provided in Appendices~\ref{app:sf-dynamic} and~\ref{app:dynamic-scheduling}, respectively.

\Needspace{12\baselineskip}
\paragraph{Self-rollout on-policy distillation.}
\label{sec:exp:onpolicy}
\begin{wraptable}{r}{0.48\textwidth}
\vspace{-8pt}
\caption{Effect of Stage 2 training.}
\label{tab:onpolicy}
\centering
\scriptsize
\setlength{\tabcolsep}{1pt}
\begin{tabular*}{\linewidth}{@{\extracolsep{\fill}}cccccc@{}}
\toprule
\multirow{2}{*}{Training} & \multirow{2}{*}{Predictor} & \multicolumn{2}{c}{81 frames}
 & \multicolumn{2}{c}{333 frames} \\
\cmidrule(lr){3-4}\cmidrule(l){5-6}
 & & Total $\uparrow$ & Speedup $\uparrow$
 & Total $\uparrow$ & Speedup $\uparrow$ \\
\midrule
\multicolumn{2}{c}{Original 4 steps} & 84.28 & 1.00$\times$
 & 82.55 & 1.00$\times$ \\
\midrule
\multirow{2}{*}{Stage 1} & DisCa & 83.86 & 1.82$\times$
 & 82.12 & 1.90$\times$ \\
 & ATC & \textbf{84.08} & 1.80$\times$
 & \textbf{82.29} & 1.89$\times$ \\
\midrule
\multirow{2}{*}{+ Stage 2} & DisCa & 84.09 & 1.82$\times$
 & 82.37 & 1.90$\times$ \\
 & ATC & \textbf{84.23} & 1.80$\times$
 & \textbf{82.71} & 1.89$\times$ \\
\bottomrule
\end{tabular*}
\end{wraptable}

Tab.~\ref{tab:onpolicy} ablates the effect of Stage~2 self-rollout on-policy distillation.
In Tab.~\ref{tab:sf-full-vbench}, DisCa uses only Stage~1 offline teacher forcing; we further apply Stage~2 to DisCa and compare both training stages for DisCa and ATC.
Stage~2 improves both predictors, while ATC retains higher Total Scores at both video lengths.
The larger gain for ATC at $333$ frames suggests that historical trajectory guidance and self-rollout training complement each other in maintaining quality over extended rollouts.
These improvements demonstrate generalization to longer videos despite training only on 81-frame sequences.

\section{Conclusion}
\label{sec:conclusion}

We presented \methodname{}, a framework for accelerating 4-step autoregressive video diffusion through historical trajectory guidance.
Our key insight is that historical features can remain informative for prediction even when they are unsuitable for direct reuse.
The ATC predictor retrieves relevant information from the previous chunk's sampling trajectory and selectively incorporates it to refine the current anchor prediction.
Self-rollout training adapts the predictor to states affected by its own approximations, while optional confidence-aware scheduling enables flexible quality--efficiency trade-offs.
Experiments on Self Forcing, Causal Forcing, and HY-WorldPlay demonstrate the effectiveness of this approach across text-to-video and camera-conditioned generation.
Notably, on Self Forcing, \methodname{} generalizes from 81-frame training videos to 333-frame generation, approximately four times the training length, achieving a $1.89\times$ denoising speedup while exceeding the original 4-step model's VBench Score.
These results support historical trajectory guidance as an effective approach to accelerating autoregressive generation over extended rollouts.
\subsection*{AI use statement}

Generative AI tools were used in this work to assist with translation,
language editing, grammar correction, and improving the clarity and
conciseness of the manuscript. They were also used to support literature
retrieval and summarization, and to assist with software development,
including code generation, modification, and debugging.

Generative AI tools were not used to develop the research methodology,
design experiments, formulate or prove mathematical claims, generate
synthetic data, or draft sections of the paper from scratch.
All AI-assisted text, literature-related information, and code were
reviewed and verified by the authors. The authors take full responsibility
for the final content of this work.

\bibliography{references}
\bibliographystyle{iclr2027_conference}

\clearpage
\appendix
\addtocontents{toc}{\protect\setcounter{tocdepth}{2}}
\begingroup
\renewcommand{\contentsname}{Appendix}
\tableofcontents
\endgroup
\clearpage
\section{Evaluation Details}
\label{app:evaluation}
\subsection{Text-to-video evaluation}
For quantitative text-to-video evaluation, we evaluate Self Forcing at $81$ and $333$ decoded frames and Causal Forcing at $81$ decoded frames, both at $832\times480$ resolution and $16$ FPS.
For text-to-video evaluation, we report VBench scores~\citep{huang2024vbench} using the LLM-extended evaluation prompts released by Self Forcing~\citep{huang2025selfforcing}.
The same extended prompts are used both to condition video generation and as the text references for text-dependent VBench scoring.

\paragraph{Full VBench evaluation.}
The main Self Forcing comparison and the ATC and self-rollout on-policy distillation ablations (Tabs.~\ref{tab:sf-full-vbench}, \ref{tab:prevcond}, and~\ref{tab:onpolicy}) evaluate all $16$ VBench dimensions.
We first normalize each dimension score using the official VBench normalization ranges.
The Quality Score is the weighted average of Subject Consistency, Background Consistency, Temporal Flickering, Motion Smoothness, Dynamic Degree, Aesthetic Quality, and Imaging Quality, assigning weight $0.5$ to Dynamic Degree and $1$ to each other dimension (total weight $6.5$).
The Semantic Score averages Object Class, Multiple Objects, Human Action, Color, Spatial Relationship, Scene, Appearance Style, Temporal Style, and Overall Consistency with equal weights.
The Total Score combines the Quality and Semantic Scores with weights of $4{:}1$.
We report all three aggregate scores for the main comparison and the Total Score for the two ablations, expressed as percentages.

\paragraph{VBench subset evaluation.}
We evaluate confidence-aware dynamic scheduling through both an ablation (Tab.~\ref{tab:conf-rollout}) and an additional comparison with competing acceleration methods on Self Forcing (Tab.~\ref{tab:sf-dynamic}).
These experiments, together with the Causal Forcing comparison in Tab.~\ref{tab:sf-dynamic}, use VBench-251, a subset of eight VBench dimensions evaluated using $251$ extended prompts, following ViDiT-Q~\citep{zhao2024vidit} and Q-VDiT~\citep{feng2025q}.
We generate one video per prompt.
Specifically, $72$ prompts from the Subject Consistency suite evaluate Subject Consistency, Motion Smoothness, and Dynamic Degree; $93$ prompts from the Overall Consistency suite evaluate Overall Consistency, Aesthetic Quality, and Imaging Quality; and $86$ prompts from the Scene suite evaluate Scene and Background Consistency.
After normalizing the dimension scores using the official VBench normalization ranges, we compute a quality score as the weighted average of Subject Consistency, Background Consistency, Motion Smoothness, Dynamic Degree, Aesthetic Quality, and Imaging Quality, assigning a weight of $0.5$ to Dynamic Degree and $1$ to each other dimension.
We average Scene and Overall Consistency to obtain a semantic score, and combine the quality and semantic scores with weights of $4{:}1$ to obtain the reported subset score as a percentage.
Subset scores are compared only within this evaluation protocol and are not directly comparable to the full VBench scores.

\subsection{Image-to-video evaluation}
HY-WorldPlay is evaluated at $125$ decoded frames at $832\times480$ resolution and 24 FPS.
For camera-conditioned HY-WorldPlay, we randomly select $50$ prompts with their corresponding initial images from the VBench I2V set, excluding all images and prompts used for training.
Following the VBench-based evaluation practice of LongLive \citep{yang2025longlive}, we evaluate a subset of five dimensions: Subject Consistency, Background Consistency, Motion Smoothness, Aesthetic Quality, and Imaging Quality, and report the arithmetic mean of their scores.
Following WorldPlay \citep{sun2025worldplay}, we use the four cyclic camera trajectories described in the Dataset subsection (Appendix~\ref{app:dataset}) to revisit previously observed views.

We additionally evaluate PSNR, SSIM, and LPIPS in two settings.
For comparison against the original 4-step model, we compare generated frames with the corresponding 4-step outputs under matched initial images, text conditions, camera trajectories, and random seeds.
For self-comparison, each model first generates frames along a camera trajectory and then retraces the same path.
We compare frames generated on the return path with the corresponding frames from the initial pass within that model's own generated sequence, following WorldPlay.
The former measures fidelity to the original model's outputs, while the latter measures consistency when revisiting the same views.

\subsection{Efficiency metrics}
All methods receive identical input conditions within each base model.
Efficiency is measured by average computation per video in TFLOPs ($10^{12}$ floating-point operations), model-path latency per video, and speedup over the corresponding full baseline.
TFLOPs and latency measure the computational cost and runtime of DiT denoising, respectively, excluding KV-cache refresh, VAE decoding, and text encoding.
Both measurements include rejected predictor attempts as well as their full-evaluation fallbacks.
Timings use matched hardware and execution conditions; speedup is the ratio of mean baseline latency to mean method latency.

\subsection{Comparison methods}
\label{app:cache}

We implement all comparison methods within the corresponding base model and autoregressive generation pipeline of Self Forcing, Causal Forcing, and HY-WorldPlay.
Within each base model, all methods use identical evaluation prompts, random seeds, and video generation settings.
At the time of our experiments, official implementations of C$^3$ache~\citep{zhao2026c3acheacceleratingworldaction} and X-Cache~\citep{zeng2026x} were not publicly available, so we do not include these methods in our empirical comparisons.
The first chunk always uses full 4-step denoising, while subsequent chunks follow each method's acceleration strategy and retain the original cross-chunk causal KV cache.
Below, F denotes a full DiT evaluation, R denotes direct cache reuse, and P denotes a learned predictor replacing a full DiT evaluation.

\paragraph{Original 4-step (\FFFF{}).}
The baseline uses the original 4-step denoising configuration, with a full DiT evaluation at every step, as the reference for generation quality and inference efficiency.
The raw timesteps are $[1000,750,500,250]$.
All configurations use the base model's timestep warping with shift $5$, $t_{\mathrm{warp}}=5t/(1+4t/1000)$, giving baseline timesteps of $[1000,937.50,833.33,625.00]$ after rounding.

\paragraph{Original 3/2-step.}
These baselines reduce the number of denoising steps in each subsequent chunk to three or two and reset the sampling timesteps accordingly.
For three-step sampling, Self Forcing and Causal Forcing use integer timesteps $[1000,666,333]$, which map to approximately $[1000,908.84,713.98]$.
HY-WorldPlay instead uses unrounded, evenly spaced timesteps $[1000,2000/3,1000/3]$, which map to approximately $[1000,909.09,714.29]$.
The two-step configuration uses $[1000,500]$ for all three bases, mapping to approximately $[1000,833.33]$.
Every retained step executes the full DiT, with no additional training, to assess the quality--efficiency trade-off of directly reducing sampling steps.

\paragraph{Naive Cache 2-step (\FRRF{}).}
This baseline is used only in the ablation study in Tab.~\ref{tab:prevcond}.
This baseline retains the original 4-step sampling schedule but evaluates the full DiT only at the first and fourth steps.
The two middle steps reuse the velocity prediction cached at the first step to perform their respective sampling updates.
Thus, each subsequent chunk still receives four denoising updates but requires only two full DiT evaluations, without feature prediction or adaptive cache decisions.

\paragraph{TeaCache.}
Our implementation of TeaCache \citep{liu2025timestep} uses the accumulated, rescaled relative change in timestep-modulated input features as a proxy for cache reuse error.
When this quantity is below a threshold, it reuses the complete DiT block stack's feature residual from the most recent full evaluation; otherwise, it executes the full stack and refreshes the cache.
The output head is evaluated at the current timestep.
To accommodate autoregressive generation with repeated timestep schedules across chunks, the change indicator uses modulated input features containing the current video content, allowing cache decisions to respond to changes in chunk content.
Base-specific calibration coefficients and cache thresholds are used, with the threshold controlling the quality--efficiency trade-off.
Each chunk independently determines its compute/reuse schedule from its own content-dependent feature changes.
This adaptation largely follows the chunk-wise independent caching principle of FlowCache \citep{ma2026flow}, so we do not include a separate FlowCache comparison; our implementation retains the TeaCache decision criterion described above.

\paragraph{TaylorSeer.}
Our implementation of TaylorSeer \citep{liu2025reusing} caches attention and feed-forward outputs within each DiT block and estimates their changes across denoising steps using finite differences at full evaluations.
Skipped computations are replaced by Taylor extrapolation of these outputs, while retaining the current timestep modulation and residual connections.
Feature histories are reset at the beginning of every chunk.
We use at most first-order prediction: two available full evaluations enable a first-order difference estimate, while a single full evaluation permits only zeroth-order reuse.
The refresh interval controls computational cost.

\paragraph{MotionCache.}
Our implementation of MotionCache \citep{xu2026motion} makes cache decisions at token granularity, estimating motion from feature differences between adjacent frames and using these estimates to weight each token's accumulated feature change.
At chunk boundaries, the previous chunk's last frame provides the preceding-frame reference.
Tokens with stronger motion or larger accumulated changes trigger recomputation earlier, while the remaining tokens reuse cached residuals.
Only selected tokens are recomputed, with their current K/V entries updated while preserving the causal attention context.
Adjusting the cache threshold controls the fraction of updated tokens and the quality--efficiency trade-off.

\paragraph{DisCa.}
We adapt the learned caching prediction approach of DisCa \citep{zou2026disca} to autoregressive video diffusion.
The predictor receives the current noisy latent and the preceding denoising step's final hidden state from the same chunk, without ATC's cross-chunk feature fusion branch.
Its DiT blocks incorporate causal KV caching to access historical video context.
The predicted final hidden state is decoded into velocity by the frozen output head.
We use the same anchor selection and update strategy as ATC, allowing the anchor to come from either a full evaluation or an accepted prediction.
For a fair comparison, we control predictor capacity and offline training conditions by using the same number of DiT blocks, training data, regression losses, and main optimization settings as ATC.
This comparison evaluates our autoregressive adaptation of DisCa rather than reproducing its original training recipe.
Depending on the target speedup, the default schedule is \FFPF{} or \FPPF{}, replacing one or two full DiT evaluations per subsequent chunk, respectively.

\section{Additional quantitative and qualitative results}
\label{app:extra}

\subsection{Additional quantitative comparisons}
\label{app:sf-dynamic}
\label{app:causal-forcing}
Tab.~\ref{tab:sf-dynamic} compares Self Forcing in the slow setting and Causal Forcing at $81$ frames using the VBench subset and efficiency metrics described in Appendix~\ref{app:evaluation}.
The step-reduction baseline is Original 3-step for Self Forcing and Original 2-step for Causal Forcing.
For Self Forcing, \methodname{} uses confidence-aware dynamic scheduling with K08, targeting an average of eight accepted predictions per video; implementation details are provided in Appendix~\ref{app:dynamic-scheduling}.
It achieves the highest VBench score among accelerated methods and the largest speedup in this setting.
For Causal Forcing, \methodname{} uses fixed \FPPF{} and achieves the highest VBench score among accelerated methods at a latency comparable to DisCa's.

\begin{table}[htbp]
\caption{Additional comparisons on 81-frame videos using the VBench subset. Ours uses dynamic K08 on Self Forcing and fixed \FPPF{} on Causal Forcing. Naive step reduction uses three and two steps, respectively.}
\label{tab:sf-dynamic}
\label{tab:causal-main}
\centering
\scriptsize
\setlength{\tabcolsep}{2.5pt}
\begin{tabular*}{\linewidth}{@{\extracolsep{\fill}}ccccc@{\hspace{8pt}}cccc@{}}
\toprule
\multirow{2}{*}{Method} & \multicolumn{4}{c}{\textbf{Self Forcing: Slow}} & \multicolumn{4}{c}{\textbf{Causal Forcing}} \\
\cmidrule(lr){2-5}\cmidrule(l){6-9}
 & VBench $\uparrow$ & TFLOPs $\downarrow$ & Latency (s) $\downarrow$ & Speedup $\uparrow$ & VBench $\uparrow$ & TFLOPs $\downarrow$ & Latency (s) $\downarrow$ & Speedup $\uparrow$ \\
\midrule
Original 4-step & 80.68 & 793 & 6.53 & 1.00$\times$ & 81.58 & 793 & 6.48 & 1.00$\times$ \\
\midrule
Original 3 / 2-step & 80.42 & 611 & 4.84 & \underline{1.35}$\times$ & 81.05 & 429 & 3.49 & \textbf{1.86$\times$} \\
TeaCache & \underline{80.47} & 611 & 4.98 & 1.31$\times$ & \underline{81.17} & 493 & 3.99 & 1.62$\times$ \\
TaylorSeer & 80.40 & 611 & 5.06 & 1.29$\times$ & 80.60 & 457 & 3.76 & 1.72$\times$ \\
MotionCache & 80.35 & 626 & 5.04 & 1.30$\times$ & 80.85 & 451 & 3.75 & 1.73$\times$ \\
DisCa & \underline{80.49} & 618 & 4.94 & \underline{1.32}$\times$ & \underline{81.28} & 443 & 3.60 & \underline{1.80$\times$} \\
\midrule
Ours & \textbf{80.65} & 559 & 4.67 & \textbf{1.40}$\times$ & \textbf{81.40} & 450 & 3.63 & \underline{1.79$\times$} \\
\bottomrule
\end{tabular*}
\end{table}

\subsection{Full VBench results on Self Forcing}
\label{app:full-vbench}
Tab.~\ref{tab:sf-vbench-dimensions} provides the $16$ individual dimension scores and three aggregate scores at $81$ and $333$ frames for all methods in Tab.~\ref{tab:sf-full-vbench}.
All dimension scores are raw values multiplied by $100$, before the normalization used for the aggregate scores.
DisCa and \methodname{} are evaluated with the fixed \FPPF{} schedule.

\begin{table}[htbp]
\caption{Per-dimension and aggregate VBench results on 81- and 333-frame Self Forcing generation.
Dimension scores are raw values multiplied by $100$; Quality, Semantic, and Total Scores are normalized weighted percentages.
Higher is better for all metrics.
Original denotes the original 4-step model.
Bold indicates the best accelerated method in each frame setting; aggregate rankings use scores before final rounding.}
\label{tab:sf-vbench-dimensions}
\centering
\scriptsize
\setlength{\tabcolsep}{2pt}
\begin{tabular*}{\linewidth}{@{\extracolsep{\fill}}cccccccc@{}}
\toprule
\multicolumn{8}{c}{\textbf{81 frames}} \\
\midrule
Dimension & Original 4-step & Original 2-step & TeaCache & TaylorSeer & MotionCache & DisCa & Ours \\
\midrule
Subject Consistency & 96.19 & 96.16 & 96.15 & 95.52 & 95.79 & 95.80 & \textbf{96.17} \\
Background Consistency & 96.96 & 96.90 & 97.01 & 95.72 & 96.88 & 96.93 & \textbf{97.22} \\
Temporal Flickering & 98.98 & 99.12 & 99.17 & 99.08 & 99.08 & 99.10 & \textbf{99.19} \\
Motion Smoothness & 98.63 & 98.58 & 98.70 & \textbf{98.72} & 98.68 & 98.58 & 98.68 \\
Dynamic Degree & 58.33 & \textbf{56.94} & 51.39 & 52.78 & 54.17 & 54.17 & 55.56 \\
Aesthetic Quality & 67.20 & 67.04 & 66.79 & 66.79 & 66.88 & 66.88 & \textbf{67.25} \\
Imaging Quality & 68.93 & 68.59 & 68.81 & \textbf{69.44} & 68.95 & 68.63 & 68.89 \\
\midrule
Object Class & 95.89 & \textbf{95.73} & 94.94 & 95.02 & 94.78 & 94.86 & 94.15 \\
Multiple Objects & 86.74 & 86.05 & 85.90 & 85.37 & \textbf{86.89} & 86.13 & 86.05 \\
Human Action & 97.00 & 96.00 & 97.00 & 96.00 & 96.00 & 96.00 & \textbf{98.00} \\
Color & 88.22 & 88.41 & 87.99 & 85.63 & 87.48 & \textbf{88.62} & 88.60 \\
Spatial Relationship & 84.60 & 84.34 & 85.60 & \textbf{86.12} & 85.23 & 84.26 & 84.94 \\
Scene & 59.08 & 58.94 & \textbf{60.32} & 59.74 & 60.10 & 59.23 & 59.01 \\
Appearance Style & 20.59 & 20.58 & 20.62 & \textbf{20.64} & 20.63 & 20.57 & 20.58 \\
Temporal Style & 26.69 & 26.59 & 26.81 & \textbf{27.04} & 26.80 & 26.71 & 26.72 \\
Overall Consistency & 27.73 & 27.69 & 27.84 & \textbf{28.01} & 27.80 & 27.80 & 27.77 \\
\midrule
Quality Score & 84.63 & 84.46 & 84.14 & 83.93 & 84.25 & 84.17 & \textbf{84.59} \\
Semantic Score & 82.87 & 82.59 & \textbf{83.01} & 82.69 & 82.85 & 82.62 & 82.80 \\
Total Score & 84.28 & 84.09 & 83.91 & 83.68 & 83.97 & 83.86 & \textbf{84.23} \\
\midrule
\addlinespace[4pt]
\multicolumn{8}{c}{\textbf{333 frames}} \\
\midrule
Dimension & Original 4-step & Original 2-step & TeaCache & TaylorSeer & MotionCache & DisCa & Ours \\
\midrule
Subject Consistency & 93.13 & \textbf{93.66} & 91.69 & 88.88 & 92.13 & 92.19 & 93.42 \\
Background Consistency & 93.81 & 94.02 & 92.67 & 90.65 & 92.50 & 93.14 & \textbf{94.13} \\
Temporal Flickering & 98.90 & \textbf{99.11} & 99.07 & 98.94 & 99.05 & 99.07 & 99.09 \\
Motion Smoothness & 98.63 & 98.62 & 98.61 & 98.61 & \textbf{98.67} & 98.58 & 98.50 \\
Dynamic Degree & 55.56 & 52.78 & 52.78 & 55.56 & 52.78 & 58.33 & \textbf{59.72} \\
Aesthetic Quality & 64.63 & \textbf{64.48} & 63.35 & 63.03 & 62.97 & 63.29 & 64.15 \\
Imaging Quality & 69.87 & \textbf{68.90} & 68.18 & 68.45 & 68.59 & 68.39 & 68.70 \\
\midrule
Object Class & 93.59 & \textbf{93.99} & 91.69 & 90.35 & 91.85 & 92.25 & 92.41 \\
Multiple Objects & 83.16 & 80.72 & 77.82 & 78.13 & 79.19 & 80.11 & \textbf{81.40} \\
Human Action & 97.00 & 96.00 & \textbf{97.00} & 95.00 & 96.00 & 96.00 & \textbf{97.00} \\
Color & 82.80 & 82.51 & 80.48 & 73.16 & 79.96 & 81.96 & \textbf{82.97} \\
Spatial Relationship & 81.91 & 82.85 & 80.33 & 80.07 & 80.19 & 82.56 & \textbf{83.13} \\
Scene & 54.94 & 54.43 & 56.47 & \textbf{58.50} & 57.85 & 57.63 & 57.05 \\
Appearance Style & 21.03 & 21.07 & 20.89 & \textbf{21.27} & 20.93 & 20.86 & 20.83 \\
Temporal Style & 26.96 & 26.89 & 27.07 & 26.85 & \textbf{27.13} & 27.09 & 26.92 \\
Overall Consistency & 27.83 & 27.86 & 28.05 & \textbf{28.13} & 27.99 & 28.08 & 27.83 \\
\midrule
Quality Score & 82.93 & 82.76 & 81.82 & 81.05 & 81.89 & 82.44 & \textbf{83.12} \\
Semantic Score & 81.04 & 80.71 & 80.06 & 79.26 & 80.24 & 80.84 & \textbf{81.06} \\
Total Score & 82.55 & 82.35 & 81.47 & 80.69 & 81.56 & 82.12 & \textbf{82.71} \\
\bottomrule
\end{tabular*}
\end{table}

\subsection{Additional qualitative results}
\label{app:qualitative}
\begin{figure}[htbp]
\centering
\includegraphics[width=\linewidth,height=0.9\textheight,keepaspectratio]{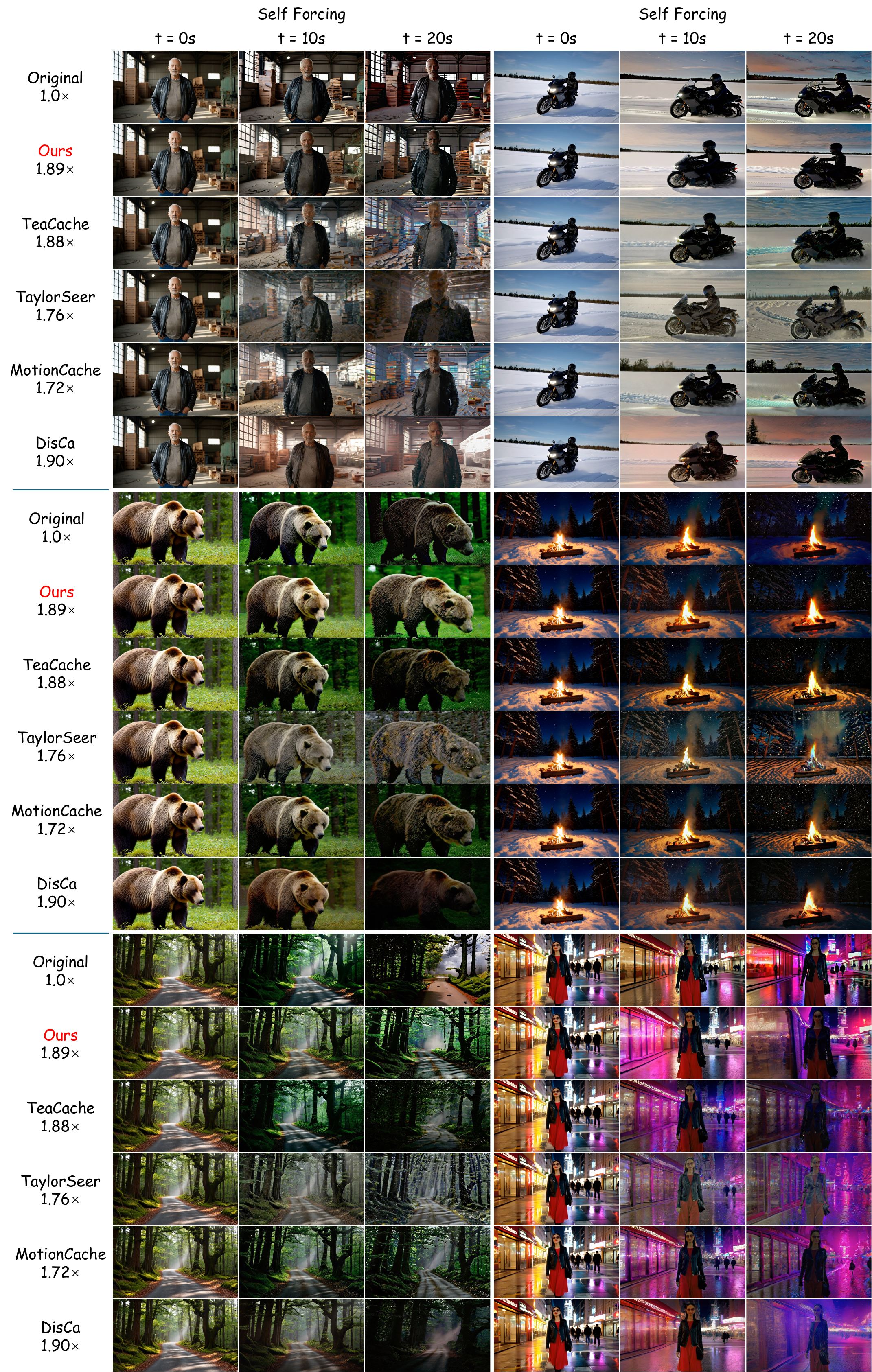}
\caption{Additional qualitative comparisons on Self Forcing.
Each example shows frames at $0$, $10$, and $20$ seconds under matched generation conditions, with denoising speedups indicated for each method.}
\label{fig:app-qualitative}
\end{figure}

Fig.~\ref{fig:app-qualitative} provides additional Self Forcing comparisons over $20$-second sequences, covering people, animals, moving vehicles, and outdoor scenes.
At a $1.89\times$ denoising speedup, \methodname{} preserves recognizable subjects and coherent scene structure in the later frames.
For example, the bear retains detailed fur and a clear body outline, while the forest path remains well defined as the viewpoint advances.
Several competing methods exhibit increasing texture artifacts, loss of detail, or appearance drift by $20$ seconds, most visibly in the workshop, bear, and nighttime street examples.
These observations complement the quantitative results at $333$ frames and support the benefit of historical trajectory guidance for maintaining visual quality over extended generation.

\section{Confidence-aware dynamic scheduling}
\label{app:dynamic-scheduling}
\paragraph{Confidence-head fitting.}
After predictor training, we freeze the generator and ATC and collect student rollouts using the deployment sampling schedule.
We save the visited states for offline confidence-head training.
At each collected state, the frozen full DiT provides a reference hidden state under the same noisy latent, timestep, and historical conditions as the ATC prediction.
The confidence head is trained from scratch to estimate the resulting local prediction error.
The head combines mean-pooled predictor features with confidence-token attention pooling, conditioned on chunk position, denoising step, and hidden-state statistics.
The head predicts log relative hidden-state error:
\begin{equation}
\epsilon=\frac{\|\hat h-h^F\|_2}{\|h^F\|_2+\delta},
\qquad \hat y=\operatorname{Conf}(z),
\qquad \hat\epsilon=\exp(\hat y),
\label{eq:conf}
\end{equation}
where $z$ collects the head inputs, $\hat h$ and $h^F$ are the ATC prediction and full-DiT reference evaluated at the same collected rollout state, and $\delta>0$ ensures numerical stability.
We train the head with the SmoothL1 objective:
\begin{equation}
\mathcal L_{\mathrm{conf}}
=\mathrm{SmoothL1}\!\left(\hat y,\log(\epsilon+\delta)\right).
\label{eq:conf-loss}
\end{equation}
Training and validation are split by prompt, and we select the checkpoint with the lowest validation loss.
For Self Forcing, training uses a global batch size of $256$ for up to $20$ epochs, with a peak learning rate of $10^{-4}$.

\paragraph{Risk estimation.}
At inference, we freeze ATC and the confidence head and select the acceptance threshold $\tau$ separately.
At each predictor-eligible step $i$ in chunk $j$, ATC first produces a candidate final hidden state $\hat h_{j,i}$ from the current noisy latent and cached history.
The confidence head then estimates its log relative error $\hat y_{j,i}$ without evaluating the full DiT.
We convert this estimate into a scheduling risk:
\begin{equation}
\hat\epsilon_{j,i}=\exp(\hat y_{j,i}),
\qquad
r_{j,i}=\hat\epsilon_{j,i}\frac{|t_i-t_{i+1}|}{T}(1+\beta\alpha_j),
\qquad
\alpha_j=\frac{K-1-j}{K-1-j_0},
\label{eq:app-scheduling-risk}
\end{equation}
where $K$ is the total number of chunks, $j_0$ is the first predictor-eligible chunk, and $t_N$ is the terminal timestep after the last denoising evaluation.
We use $T=1000$ and $\beta=2$.
The outgoing denoising span accounts for the sampling update affected by the prediction, while $\alpha_j$ decreases from one at $j_0$ to zero at the final chunk to assign greater weight to errors that can propagate through more subsequent chunks.

\paragraph{Acceptance and full-DiT fallback.}
If $r_{j,i}\leq\tau$, we accept the candidate hidden state.
Otherwise, we discard the candidate and evaluate the full DiT on the same noisy latent, timestep, and conditions before making any sampling update:
\begin{equation}
h_{j,i}=
\begin{cases}
\hat h_{j,i}, & r_{j,i}\leq\tau,\\
\operatorname{DiT}_{\theta}^{\mathrm{final}}(x_i^{j},t_i,\mathcal{C}^{j}), & r_{j,i}>\tau,
\end{cases}
\label{eq:routing}
\end{equation}
where $\operatorname{DiT}_{\theta}^{\mathrm{final}}$ denotes the final hidden state produced by the full DiT, and $\mathcal{C}^{j}$ contains the external conditions and historical KV context.
Thus, the full DiT is evaluated only when the candidate is rejected at a predictor-eligible step.
Steps that are not predictor-eligible directly use the full DiT.

\paragraph{Sampling and cache updates.}
The selected hidden state is decoded through the frozen output head and used by the base model's sampler:
\begin{equation}
v_{j,i}=\mathrm{Head}(h_{j,i},t_i),
\qquad
x_{i+1}^{j}=\mathcal S_i\!\left(x_i^{j},v_{j,i}\right).
\label{eq:app-scheduled-update}
\end{equation}
We also cache $h_{j,i}$ for subsequent predictions, including same-chunk anchoring and historical guidance at the matching timestep of the next chunk.
After a fallback, this cache therefore contains the full-DiT hidden state rather than the rejected prediction.
Both accepted and fallback outputs can consequently become references for later ATC calls.
Because ATC and the confidence head run before the acceptance decision, rejected attempts incur their computation in addition to the full-DiT evaluation; all reported efficiency measurements include this cost.

\paragraph{Scheduling ablation.}
Tab.~\ref{tab:conf-rollout} compares fixed schedules and confidence aware dynamic scheduling on 81-frame Self Forcing videos.
K06, K08, and K10 denote target average counts of $6$, $8$, and $10$ accepted predictions per video, respectively.
The dynamic configurations use the same predictor and confidence head, varying only the acceptance threshold.
K08 improves generation quality over both \FPFF{} and \FFPF{} while reducing latency, whereas K10 provides a faster operating point with a modest quality tradeoff.
Compared with fixed \FPPF{} under the same VBench subset evaluation, K08 provides higher generation quality at a lower acceleration level.
This ablation complements the comparisons with other acceleration methods in Appendix~\ref{app:sf-dynamic}.

\begin{table}[htbp]
\centering
\caption{Dynamic schedule ablation on 81-frame Self Forcing videos, evaluated on the VBench subset.}
\label{tab:conf-rollout}
\centering
\scriptsize
\setlength{\tabcolsep}{1.5pt}
\renewcommand{\arraystretch}{1.05}
\begin{tabular}{ccccc}
\toprule
Method & VBench $\uparrow$ & TFLOPs $\downarrow$ & Latency (s) $\downarrow$ & Speedup $\uparrow$ \\
\midrule
\shortstack[c]{Original 4 steps} & 80.68 & 793 & 6.53 & 1.00$\times$ \\
\midrule
\FPFF{} & 80.55 & 622 & 4.99 & 1.31$\times$ \\
\FFPF{} & 80.57 & 622 & 4.99 & 1.31$\times$ \\
Dynamic K06 & \underline{80.62} & 613 & 5.10 & 1.28$\times$ \\
Dynamic K08 & \textbf{80.65} & 559 & 4.67 & \underline{1.40}$\times$ \\
Dynamic K10 & \underline{80.58} & 507 & 4.24 & \underline{1.54}$\times$ \\
\FPPF{} & 80.55 & 452 & 3.62 & \textbf{1.80}$\times$ \\
\bottomrule
\end{tabular}
\end{table}

\section{Training details}
\label{app:hparams}
\label{app:setup-details}

\subsection{Stage 1: offline feature regression}
\paragraph{Shared supervision.}
Each predictor block uses the historical and text attention K/V from its corresponding source layer.
Supervision targets the final hidden states of the complete teacher DiT and their corresponding velocities, rather than the intermediate outputs of the selected layers.
Both the within-chunk anchor and the previous chunk's same-timestep reference are taken from offline teacher trajectories.
The denoising residual output and cross-chunk correction output are zero-initialized, so the prediction initially copies the anchor.
The first chunk provides context only, while each subsequent chunk contributes three prediction positions.
We optimize a combination of hidden-state and velocity MSE:
\begin{equation}
\mathcal L_{\mathrm{S1}}
=\lambda_h\,\mathrm{MSE}(\hat h,h^F)
+\lambda_v\,\mathrm{MSE}(\mathrm{Head}(\hat h,t_i),v^F).
\label{eq:stage1}
\end{equation}
Here $h^F$ and $v^F$ are the final hidden state and velocity from the full DiT evaluated at the same input state.
We set $\lambda_h=0.1$ and $\lambda_v=1$, without additional timestep weighting or relative-error normalization.

\paragraph{Self Forcing and Causal Forcing.}
ATC uses a single predictor block initialized from layer $17$ of the corresponding pretrained generator.
The source-layer selection study for Self Forcing is provided in Appendix~\ref{app:block-sweep}.
We collect 81-frame teacher rollouts for offline supervision.
Training cycles through shuffled chunk--timestep groups; each update samples $64$ distinct prompts from one group, keeping the historical context length consistent within a batch.
We train for $2000$ updates with a global batch size of $64$ using AdamW, with betas of $(0.9,0.95)$, weight decay of $0.01$, and gradient clipping at $1.0$.
The learning rates are $10^{-4}$ for the input/fusion parameter group and $10^{-5}$ for the predictor block.
We use cosine learning-rate decay.
The final checkpoint initializes Stage~2.

\paragraph{HY-WorldPlay.}
ATC uses two predictor blocks initialized from layers $1$ and $52$ of the pretrained generator, with zero-based indexing.
Offline supervision uses 125-frame teacher rollouts conditioned on the initial image, text prompt, and camera trajectory, with the shared supervision protocol and loss defined above.
We train for $2000$ updates with an effective batch size of $64$, using AdamW with betas $(0.9,0.95)$, weight decay $0.01$, and gradient clipping at $1.0$.
The learning rates are $10^{-4}$ for the fusion module and residual outputs and $10^{-5}$ for the two predictor blocks, with cosine decay.

\subsection{Stage 2: self-rollout on-policy distillation}
\label{app:onpolicy-training}
\textbf{Student rollout and random exit.}
Self Forcing and Causal Forcing use the same Stage~2 training procedure, initialized from the final Stage~1 checkpoint.
Each rollout generates an 81-frame video with the base generator frozen.
For each predictor update, we randomly choose an exit step $i_{\mathrm{exit}}\in\{1,2,3\}$, shared across chunks.
The first chunk uses full DiT evaluations up to the selected exit.
For subsequent chunks, the first step uses the full DiT, and ATC drives the remaining sampling steps and updates the feature cache.
Each chunk exits with its predicted clean latent, which provides context for the next chunk.
Varying the exit exposes ATC to different prediction depths and to historical features produced by its own earlier predictions.
Gradients are retained only for the exit-step ATC predictions; earlier rollout states, anchor features, historical features, and KV caches are detached.

\textbf{DMD supervision.}
Following Self Forcing~\citep{huang2025selfforcing}, we use a frozen non-causal Wan2.1-T2V-14B real-score teacher and a trainable non-causal Wan2.1-T2V-1.3B fake-score critic for DMD~\citep{yin2024one,yin2024improved} supervision of the generated video.
The score networks evaluate the complete video, while the ATC objective excludes the first chunk and adds no Stage~1 reconstruction loss.
The critic is trained with flow matching on detached student samples, including all chunks.

\textbf{Training parameters.}
We perform $2000$ predictor updates with a global batch size of $8$, with five critic updates on fresh samples before each predictor update.
AdamW uses betas $(0,0.999)$, weight decay $0.01$, and learning rates $10^{-5}$, $10^{-6}$, and $10^{-5}$ for the ATC fusion module, predictor block, and critic, respectively.
We use cosine learning-rate decay.
Predictor EMA starts at update $200$ with decay $0.99$.
The DMD score timestep is shared across the video, sampled from integer timesteps in $[0,1000)$, warped with shift $5$, and clipped to $[20,980]$.
Teacher guidance uses $D_{\mathrm{cond}}+3(D_{\mathrm{cond}}-D_{\mathrm{uncond}})$.

\textbf{HY-WorldPlay.}
For training efficiency, we fine-tune HY-WorldPlay's predictor using direct teacher supervision on student rollouts, avoiding the additional score-network training and alternating updates required by DMD.
We initialize ATC from Stage~1 and keep the full teacher frozen.
Each training rollout generates 125 frames.
The first chunk uses the full DiT throughout; subsequent chunks use the full DiT at the first denoising step and ATC at the remaining steps.
Predicted velocities advance the noisy latents, and the resulting hidden states and generated chunks provide the inputs and historical context for later predictions.

At each predictor step, the full teacher evaluates the same noisy latent reached by the student and supplies the target final hidden state $h^F$ and velocity $v^F$.
We minimize
\begin{equation}
\mathcal L_{\mathrm{HY}}
=\mathbb E\!\left[
\lambda_h\,\operatorname{MSE}(\hat h,h^F)
+\lambda_v\,\operatorname{MSE}(\hat v,v^F)
\right].
\label{eq:app-hy-rollout-loss}
\end{equation}
The expectation averages over predictor steps on student rollouts, using the same loss weights as Stage~1.
The teacher outputs serve only as supervision; the student outputs drive subsequent sampling updates.
Rollout states and cached features are detached between denoising steps and chunks, so each prediction receives local supervision without backpropagation through the complete trajectory.

We train for $2000$ updates with an effective batch size of eight trajectories, using learning rates of $5\times10^{-5}$ for the fusion module and $5\times10^{-6}$ for the predictor block.
Weight decay and gradient clipping are set to $0.01$ and $1.0$, respectively.

\subsection{Dataset}
\label{app:dataset}
\textbf{Self Forcing and Causal Forcing.}
We sample training prompts from VidProM \citep{wang2024vidprom}.
Stage~1 uses offline trajectories generated by the corresponding full model, with final hidden states and velocities providing supervision for the predictor.
Stage~2 samples prompts from the filtered, extended VidProM pool and generates training trajectories online with the current predictor.

\textbf{HY-WorldPlay.}
Both training stages use randomly selected prompts and their corresponding initial images from the VBench I2V set.
The image--prompt pairs used for training are disjoint from those used for evaluation, with no overlap in either input images or prompts.
We combine these image--prompt pairs with camera trajectories that move forward then backward, move left then right, tilt up then down, or pan left then right.
Stage~1 uses offline full-teacher trajectories, while Stage~2 generates student rollouts under the same image, prompt, and camera conditions and obtains teacher supervision at the visited states.

\section{Details of the Motivating Experiments}
\label{app:motivation-diagnostics}

\paragraph{Correlation along denoising time and video time.}
Fig.~\ref{fig:boundary}(a) measures token cosine similarity of final hidden states between consecutive denoising timesteps and between the previous chunk's last temporal slot and the current chunk at the same timestep.
The $50$-step references use a bidirectional DiT teacher for Self Forcing and a causal DiT teacher for Causal Forcing.
Spatial alignment uses optical flow from generated RGB frames; hatched bars show the aligned results.
Alignment increases the 4-step cross-chunk mean by $0.0138$/$0.0164$ for Self Forcing/Causal Forcing. Optical-flow alignment is used only for this diagnostic and is not part of ATC inference.

\paragraph{Global impact of local approximation.}
Fig.~\ref{fig:boundary}(b) applies \FRRF{} to one chunk at a time in a 165-frame rollout, keeping every other chunk \FFFF{}.
Each R reuses that chunk's first full-step velocity.
Curves average full-video PSNR, SSIM, and LPIPS over the first $10$ MovieGenVideoBench-extended prompts (seed $0$), relative to matched \FFFF{} references. These full-video metrics assess the overall impact of approximation placement, including the different lengths of the remaining rollout.

\section{Predictor block selection on Self Forcing}
\label{app:block-sweep}

We study source-layer selection and initialization using a single-block predictor.
Candidates are trained under the same offline regression protocol and compared by validation velocity MSE on held-out teacher states.
Each source layer supplies both the initial block weights and the corresponding attention K/V; layer indices are zero-based.

\subsection{Source layer and initialization}

Tab.~\ref{tab:block-source} reports all $30$ source layers.
Layer $17$ has the lowest validation velocity MSE ($0.046459$) among the single-block candidates.
These results motivate using layer $17$ for subsequent Self Forcing predictor training.

\begin{table}[t]
\caption{Validation velocity MSE across all source layers.}
\label{tab:block-source}
\centering
\small
\setlength{\tabcolsep}{6pt}
\begin{tabular*}{\linewidth}{@{\extracolsep{\fill}}cccccc@{}}
\toprule
Source layer & Val. MSE $\downarrow$ & Source layer & Val. MSE $\downarrow$ & Source layer & Val. MSE $\downarrow$ \\
\midrule
0 & 0.052675 & 10 & 0.049378 & 20 & 0.047325 \\
1 & 0.051242 & 11 & 0.048483 & 21 & 0.046994 \\
2 & 0.048375 & 12 & 0.049627 & 22 & 0.058777 \\
3 & 0.048407 & 13 & 0.048251 & 23 & 0.048755 \\
4 & 0.048115 & 14 & 0.049588 & 24 & 0.047499 \\
5 & 0.048193 & 15 & 0.049718 & 25 & 0.048287 \\
6 & 0.047933 & 16 & 0.047317 & 26 & 0.048326 \\
7 & 0.049241 & 17 & \textbf{0.046459} & 27 & 0.049492 \\
8 & 0.048467 & 18 & 0.047587 & 28 & 0.051012 \\
9 & 0.049020 & 19 & 0.047851 & 29 & 0.050500 \\
\bottomrule
\end{tabular*}
\end{table}

Four additional controls keep layer-$17$ K/V fixed and change only block initialization (Tab.~\ref{tab:block-init}).
Initializing from the complete teacher block yields the lowest validation velocity MSE among the tested initialization schemes under the same training budget.

\begin{table}[t]
\caption{Initialization controls with layer-17 K/V fixed, evaluated by validation velocity MSE.}
\label{tab:block-init}
\centering
\small
\setlength{\tabcolsep}{6pt}
\begin{tabular}{cc}
\toprule
Initialization & Val. MSE $\downarrow$ \\
\midrule
Complete teacher block & \textbf{0.046459} \\
Teacher + zeroed branch outputs & 0.055906 \\
Xavier + zeroed branch outputs & 0.057737 \\
Xavier & 0.057085 \\
All-zero block & 0.060460 \\
\bottomrule
\end{tabular}
\end{table}

\end{document}